\documentclass[mathpazo]{csam}

\usepackage{tabularx}
\usepackage{color} 
\usepackage{diagbox}
\usepackage{graphics}
\usepackage{multirow}
\usepackage{bm}
\usepackage{float} 
\usepackage{amsmath}
\usepackage{epsfig}
\usepackage{amssymb,amsfonts}
\usepackage{epstopdf}
\usepackage{tikz}
\usepackage{pgflibraryarrows}
\usepackage{pgflibrarysnakes}
\usepackage{mlmath}
\usepackage{natbib} 
\usepackage{subfig}
\usepackage{url}

\usepackage[margin=1.5in]{geometry}    
\usepackage[utf8]{inputenc}
\usepackage{algorithm}
\usepackage{arevmath}     
\usepackage[noend]{algpseudocode}

\makeatletter
\newcommand*\bigcdot{\mathpalette\bigcdot@{.5}}
\newcommand*\bigcdot@[2]{\mathbin{\vcenter{\hbox{\scalebox{#2}{$\m@th#1\bullet$}}}}}
\makeatother

\usepackage{changepage}

\begin{document}
\title{Frequency Principle in Deep Learning Beyond Gradient-descent-based Training}


 \author[Y. Ma et~al.]{Yuheng Ma\affil{1}\comma,
       Zhi-Qin John Xu\affil{2}\corrauth~and Jiwei Zhang\affil{1*}}
 \address{\affilnum{1}\ School of Mathematics and Statistics, and Hubei Key Laboratory of Computational Science, Wuhan University, Wuhan 430072, P.R. China. \\
           \affilnum{2}\ School of Mathematical Sciences, Institute of Natural Sciences, MOE-LSC and Qing Yuan Research Institute,
    Shanghai Jiao Tong University, Shanghai 200240, P.R. China.
    }
 \emails{ {\tt xuzhiqin@sjtu.edu.cn} (Z.~Xu),
          {\tt jiweizhang@whu.edu.cn} (J.~Zhang)}



\begin{abstract}
Frequency perspective recently makes progress in understanding deep learning. It has been widely verified in both empirical and theoretical studies that deep neural networks (DNNs) often fit the target function from low to high frequency, namely Frequency Principle (F-Principle). F-Principle sheds light on the strength and the weakness of DNNs and inspires a series of subsequent works, including theoretical studies, empirical studies and the design of efficient DNN structures etc. Previous works examine the F-Principle in gradient-descent-based training. It remains unclear whether gradient-descent-based training is a necessary condition for the F-Principle. In this paper, we show that the F-Principle exists stably in the training process of DNNs with non-gradient-descent-based training, including optimization algorithms with gradient information, such as conjugate gradient and BFGS, and algorithms without gradient information, such as Powell's method and Particle Swarm Optimization. These empirical studies show the universality of the F-Principle and provide hints for further study of F-Principle.
\end{abstract}


\maketitle


\section{Introduction}

Understanding deep neural networks (DNNs) is an important problem in modern machine learning, since it has permeated many aspects of daily life and important industries.  Recent studies find that DNNs with gradient-descent-based algorithms often follow a
Frequency Principle (F-Principle) proposed in \cite{xu_training_2018,xu2019frequency} or \cite{rahaman2018spectral}, namely,
\begin{changemargin}{0.5cm}{0.5cm}\emph{DNNs tends to learn target
functions from low frequency to high frequency during the training.} \end{changemargin}

Theoretical studies subsequently  show  that frequency principle holds in general setting with infinite samples \citep{luo2019theory} and in the regime of wide neural networks (Neural Tangent Kernel (NTK) regime \citep{jacot2018neural}) with finite samples \citep{zhang2019explicitizing,luo2020exact,luo2020fourier} or sufficient many samples \citep{cao2019towards,yang2019fine,basri2019convergence,bordelon2020spectrum}. \cite{e2019machine} show that the integral equation would naturally  lead  to the frequency principle. With the theoretical understanding, the frequency principle inspires the design of DNN-based algorithms \citep{liu2020multi,wang2020multi,you2019drawing,jagtap2019adaptive,cai2019phasednn,biland2019frequency,li2020multi}. In addition, the F-Principle provides a mechanism to understand many phenomena in applications and inspires a series of study on deep learning from frequency perspective, such as the generalization of DNNs \citep{xu2019frequency,ma2020slow}, the understanding of the effect of depth in DNNs \citep{xu2020deep}, the difference between the traditional algorithm and DNN-based algorithm in solving PDEs \citep{wang2020implicit}. 

All previous studies of the F-Principle consider the gradient-descent-based training. It is still unclear whether the gradient-descent-based training is a necessary condition for the F-Principle in DNN training process. Previous studies of the F-Principle in finite-width network \citep{xu2019frequency,luo2019theory} or infinite width network\citep{zhang2019explicitizing,basri2019convergence,yang2019fine,cao2019towards} all base  on the gradient flow of the training. However,  the deep learning can be trained by many optimization algorithms in addition to the (stochastic) gradient descent. In this work, we use numerical experiments to show that the F-Principle holds stably in the DNN training process with non-gradient-descent algorithms, such as conjugate gradient and BFGS. We further show that the F-Principle can also  exist in the DNN training process with optimization algorithms without any gradient information in each iteration step, such as Powell's method and Particle Swarm Optimization. To further show the universality of the F-Principle, we design an Monte-Carlo-Like  optimization algorithm  that randomly selects parameters, which can decrease the loss function. During the training of this Monte-Carlo-Like optimization algorithm, we found that the F-Principle still holds well. 

The rest of paper is organized as follows. Section \ref{sec:expdetail} introduces experimental details. Section \ref{sec:gdmethod} shows that gradient descent is not necessary for the F-Principle. Section \ref{sec:gnot} shows that gradient information during training is not necessary for the F-Principle. A short discussion and conclusion are  given in section\ref{sec:conclusion}. 

\section{Experimental details}\label{sec:expdetail}
To examine the F-Principle, it requires to differentiate the low- and high-frequency parts of dataset. In the following, we introduce discrete Fourier transform for 1d synthetic data and a filtering method for high-dimensional dataset, proposed in \cite{xu2019frequency}.
\subsection{Discrete Fourier transforms in synthetic data}
Following the suggested notation in \cite{beijing2020Suggested}, we denote the target function by $f(x)$ and the DNN output by  $f_{\theta}(x)$.  
For $1$-d synthetic data, $\{(x_j,f(x_j))\}_{j=1}^n$ and  $\{(x_j,f_\theta(x_j))\}_{j=1}^n$, we examine the  relative error of each frequency during the training. The discrete Fourier transforms (DFT) of $f(x)$
and the DNN output (denoted by $f_{\theta}(x)$) are computed by: 
$$\hat{f}_{k}=\frac{1}{n}\sum_{j=1}^{n}f(x_{j})\E^{-\I2\pi jk/n}, \text{\;\;and \;\;}\hat{f_{\theta}}_{k}=\frac{1}{n}\sum_{j=1}^{n}f_{\theta}(x_{j})\E^{-\I2\pi jk/n},$$
where $k$ is the frequency. We compute the relative difference between the DNN output
and the target function for each frequencies $k$
at each training epoch, that is, $\Delta_{F}(k)=|\hat{f_{\theta}}_{k}-\hat{f}_{k}|/|\hat{f}_{k}|$, where $|\cdot|$ denotes the norm of a complex number. 

F-Principle is tenable when $\Delta_F(k)$ converge to $0$ one by one in the training process, from low frequency components to high-frequency components. For our experiments of 1-d situation in the following texts, $\{x_j\}^{n}_{j=1}$ will choose evenly sampled points from $[-3.14,3.14]$ with sample size 201, and each elements in $\bm{W}^{[l]}$ and $\bm{b}^{[l]}$ are initialized by a distribution, namely, they are sampled from $N(0, \frac{2}{m_{l+1}+m_l})$, where $m_l$ is the neuron number of $l$th layer. The loss function is  chosen  as mean squared error (MSE)  here. The activation function for fully-connected networks is sigmoid function.

\subsection{Real data}
For high-dimensional data, it is hard to compute the high-dimensional Fourier transform. We now introduce a filtering method proposed in \cite{xu2019frequency} to examine the F-Principle in a real data set (e.g., MNIST).

We train the DNN by the \emph{original dataset}
$\{(\bm{x}_{i}, \bm{y}_{i})\}_{i=0}^{n-1}$, where $\bm{x}_{i}$ is an image vector, $\bm{y}_{i}$ is a one-hot vector. At each training epoch, the low frequency part can be derived
by a low-frequency filter, that is, the convolution with a Gaussian function, 
\begin{equation}
    \bm{y}_{i}^{{\rm low},\delta}=\frac{1}{C_{i}}\sum_{j=0}^{n-1}\bm{y}_{j}G^{\delta}(\bm{x}_{i}-\bm{x}_{j}),\label{eq:filter}
\end{equation}
where $C_{i}=\sum_{j=0}^{n-1}G^{\delta}(\bm{x}_{i}-\bm{x}_{j})$
is a normalization factor and $\delta$ is the variance of the following Gaussian function 
\begin{equation}
    G^{\delta}(\bm{x}_{i}-\bm{x}_{j})=\exp\left(-|\bm{x}_{i}-\bm{x}_{j}|^{2}/(2\delta)\right).
\end{equation}
The high frequency part can be derived by 
$$\bm{y}_{i}^{\mathrm{high},\delta}\triangleq\bm{y}_{i}-\bm{y}_{i}^{\mathrm{low},\delta}.$$
We also compute $\bm{h}_{i}^{\mathrm{low},\delta}$ and $\bm{h}_{i}^{\mathrm{high},\delta}$ for each DNN output $\bm{h}_{i}=f_{\theta}(x_i)$.

To quantify the convergence of $\bm{h}^{\mathrm{low},\delta}$
and $\bm{h}^{\mathrm{high},\delta}$, we compute the relative error
$e_{\mathrm{low}}$ and $e_{\mathrm{high}}$ at each training epoch,
\begin{align}
  &  e_{\mathrm{low}}=\left(\frac{\sum_{i}|\bm{y}_{i}^{\mathrm{low},\delta}-\bm{h}_{i}^{\mathrm{low},\delta}|^{2}}{\sum_{i}|\bm{y}_{i}^{\mathrm{low},\delta}|^{2}}\right)^{\frac{1}{2}},\\
  &   e_{\mathrm{high}}=\left(\frac{\sum_{i}|\bm{y}_{i}^{\mathrm{high},\delta}-\bm{h}_{i}^{\mathrm{high},\delta}|^{2}}{\sum_{i}|\bm{y}_{i}^{\mathrm{high},\delta}|^{2}}\right)^{\frac{1}{2}},\label{eq:ehigh}
\end{align}
where $\bm{h}^{\mathrm{low},\delta}$ and $\bm{h}^{\mathrm{high},\delta}$
are obtained from the DNN output, which evolves as a function of training epoch, through the same decomposition.
If $e_{\mathrm{low}}<e_{\mathrm{high}}$ for different $\delta$'s during
the training, F-Principle holds; otherwise, it is falsified. 

We use a MSE loss and a small sigmoid-CNN network, i.e., two convolutional layers (one convolution layer of 5$\times$5$\times$32, a max pooling of 2$\times$2, one convolution layer of 5$\times$5$\times$64, a pooling layer of 2$\times$2), followed by a fully connected multi-layer neural network 1024-10 equipped with a softmax.

Due to the memory constrained of some training algorithms, we only train 550 randomly selected samples from MNIST data, and we only perform experiments of Conjugate Gradient algorithm and L-BFGS in the following experiments. Other algorithms perform badly for the high-dimensional MNIST data.

\section{Gradient descent is not necessary for F-Principle} \label{sec:gdmethod}
In this section, we would examine the F-Principle in the DNN training with algorithms which are non-gradient-descent algorithms but still use gradient information in each iteration step.

The algorithms used in this section are variants of Newton method. Newton method is faster than the Gradient Descent in terms of iteration step number. However, due to the difficulty of ensuring Hessian matrix positive definite and the complexity of computing the inverse of the Hessian matrix, the original Newton's method is rarely used for large scale computations. Instead, the conjugate gradient algorithm (CG), truncated Newton algorithm (TNC), BFGS \citep{Wright06} and its variant L-BFGS is popular for practical simulations.

\subsection{Conjugate Gradient algorithm }

Conjugate gradient (CG) algorithm is a popular algorithm for solving nonlinear optimization problems. The features of CG are that it requires no matrix storage and are faster than the gradient descent. The detail of CG algorithm is given in Appendix.

\begin{figure}[h]
\centering
\includegraphics[width=2.7in]{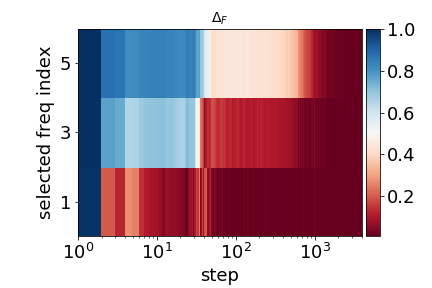}
\caption{Using CG to learn $f(x)= \sin x+\sin 3x+\sin 5x $. $\Delta_{F}(k)$
of three selected important frequencies against different training epochs. Blue indicates large relative error, while red indicates small relative error.  }
\label{fig:cg}
\end{figure}

We use CG algorithm to train the 1-100-10-1 DNN to learn the target function with three frequency peaks, namely,
\begin{equation}\label{TF3}
f(x)= \sin x + \sin 3x + \sin 5x.
\end{equation}
 As shown in Figure \ref{fig:cg}, the DNN converges gradually from low-frequency to high frequency in Fourier analysis.

\begin{figure} 
    \begin{centering}
    \subfloat[$\delta=2$]{
        \includegraphics[width=2.5in]{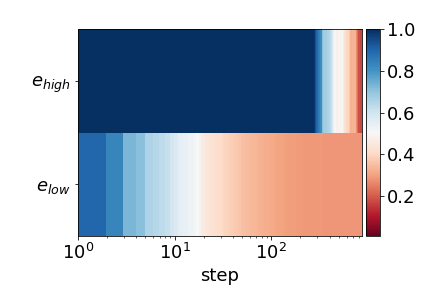}
    }
    \subfloat[$\delta=7$]{
        \includegraphics[width=2.5in]{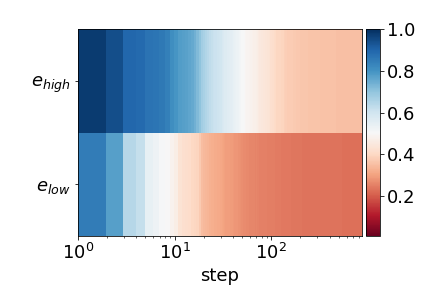}
    }
    \par\end{centering}
    \caption{Using CG to learn MNSIT. The F-Principle is examined for different $\delta$. $e_{\mathrm{low}}$ and $e_{\mathrm{high}}$ indicated by color against training epoch. }\label{fig:cgmnist}
\end{figure}

 We also use CG algorithm to train a convolutional network to fit MNIST. The F-Principle also holds in the training process, see Figure  \ref{fig:cgmnist} for  different variances $\delta$ of the Gaussian function.

\subsection{Truncated Newton algorithm}

Truncated Newton algorithm (TNC), also called Newton Conjugate-Gradient, is a nonlinear method   based on  Newton method. The  CG method is designed to solve positive definite systems, however,  the Hessian matrix may have negative eigenvalues to lead to an inaccurate solution.  TNC is a Hessian-free optimization method \citep{Nash84}. The detail of TNC algorithm is given in Appendix.  

\begin{figure}[h]
\centering
\includegraphics[width=2.7in]{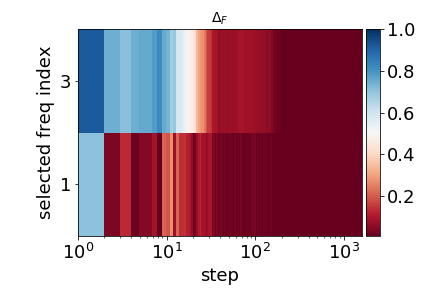}
\caption{Using TNC to learn $f(x)= \sin x+\sin 3x$. $\Delta_{F}(k)$
of three selected important frequencies against different training epochs. }
\label{fig4:tnc}
\end{figure}

In this experiment, we use TNC to train the 1-100-10-1 DNN to learn the target function with two frequency peaks, i.e., $f(x)= \sin x + \sin 3x $. Again, as shown in Figure \ref{fig4:tnc}, the DNN converges gradually from low-frequency to high frequency.

In our experiments, we point out that the results of training DNN by TNC to regress the target function \eqref{TF3} are not perfect, which is caused by the fact that the search direction is not the descent direction. This also may be influenced by the Hessian matrix, which may not keep positive definite in the training process. For $f(x)=\sin x+\sin 3x$, the low-frequency components is converged first, and it is regressed perfectly. 

\subsection{BFGS and L-BFGS}

In this subsection, we use BFGS and L-BFGS, which are quasi-Newton methods, to train neural networks. The detail of BFGS algorithm is given in Appendix.

In Figure \ref{fig:lbfgs}(a), we use BFGS to train a DNN of 1-100-10-1  to learn  the target function \eqref{TF3}. Similarly, the DNN converges gradually from low-frequency to high frequency.

\begin{figure} 
    \begin{centering}
    \subfloat[BFGS]{
        \includegraphics[width=2.5in]{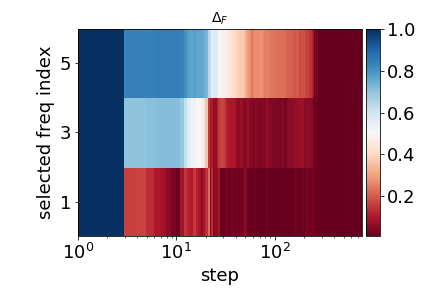}
    }
    \subfloat[L-BFGS]{
        \includegraphics[width=2.5in]{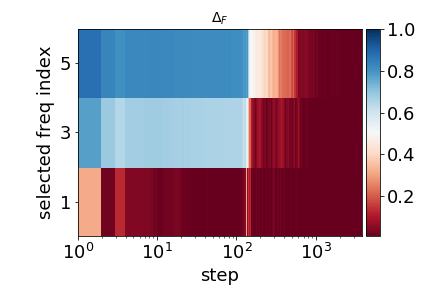}
    }
    \par\end{centering}
    \caption{Using BFGS in (a) and L-BFGS in (b) to learn $f(x)= \sin x + \sin 3x +\sin 5x $.  $\Delta_{F}(k)$
of three selected important frequencies against different training epochs. }\label{fig:lbfgs}
\end{figure}

L-BFGS uses a limited memory algorithm based on BFGS. This method only uses the data of the recent steps, which can simplify the  computation and memory \citep{Wright06,Byrd95}. The detail of L-BFGS algorithm  is  given in Appendix.

In Figure \ref{fig:lbfgs}(b), we use L-BFGS to train a DNN of 1-500-50-1  to learn the target function \eqref{TF3}. It is clear that the DNN converges gradually from low-frequency to high frequency in this example.

\begin{figure} 
    \begin{centering}
    \subfloat[$\delta=2$]{
        \includegraphics[width=2.5in]{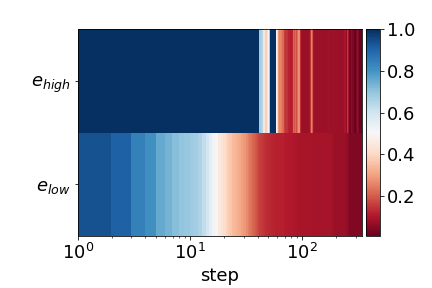}
    }
    \subfloat[$\delta=7$]{
        \includegraphics[width=2.5in]{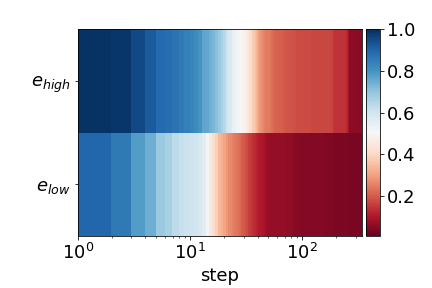}
    }
    \par\end{centering}
    \caption{Using L-BFGS to learn MNSIT. The F-Principle is examined for different $\delta$.  $e_{\mathrm{low}}$ and $e_{\mathrm{high}}$ indicated by color against training epoch. }\label{fig:lbfgsmnist}
\end{figure}

We further use L-BFGS to train a CNN to learn MNIST. As shown in Fig. \ref{fig:lbfgsmnist}, for different filter width, we still observe that the high frequency part converges slower, that is, F-Principle.

In this section, we have used experiments to show that a training algorithm, which uses gradient information but not a gradient-descent method, can still lead to the phenomenon of F-Principle. In the next section, we would show that even using a training algorithm without using gradient information, the F-Principle can still  hold.
  
\section{Gradient is not necessary} \label{sec:gnot}
In this section, we use two non-gradient-based optimization algorithm  (i.e., Powell's method,  Particle Swarm Optimization (PSO)) and a Monte-Carlo-like algorithm, to examine the F-Principle. 

\subsection{Powell's method}

Powell's method, a conjugate direction method, performs sequential one-dimensional minimization  along each vector of a direction set, in  which the direction set is updated at each iteration \citep{powell1964efficient}. The method  used here is a modification of Powell's method. The loss function can be non-differentiable, since no derivative is taken. The detail of Powell's method is given in Appendix.

\begin{figure} 
    \begin{centering}
    \subfloat[Powell's Method]{
        \includegraphics[width=2.5in]{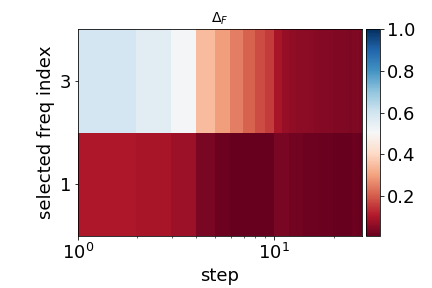}
    }
    \subfloat[PSO]{
        \includegraphics[width=2.5in]{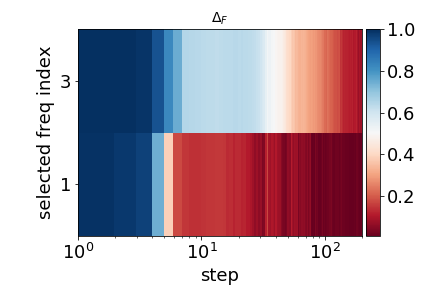}
    }
    \par\end{centering}
    \caption{Using non-gradient-based method to learn $f(x)= \sin x + \sin 3x $. $\Delta_{F}(k)$
of selected important frequencies against different training epochs.}\label{fig:nongrad}
\end{figure}

 The Powell's method is slow in solving large scale optimization problems, such as training DNN, since it needs large internal memory and computations. Therefore, we use Powell's method to train a small DNN of 1-100-1 to learn $f(x)= \sin x + \sin 3x $. Considering the two frequency peaks of the target function, as shown in Figure \ref{fig:nongrad}(a), one  can see that the DNN converges the low-frequency components first.


\subsection{Particle Swarm Optimization}

 Particle Swarm Optimization (PSO) does random search in parameter space, inspired by the moving of the swarm of birds. It can be viewed as a mid-level form of artificial life or biologically derived algorithm, and  highly depends on stochastic processes.  The PSO algorithm is given in appendix:

In this experiment, we use PSO to train a DNN of 1-100-10-1 to learn target function: $f(x)= \sin x + \sin 3x $. As shown in Figure \ref{fig:nongrad}(b), the DNN learns the low-frequency components first.


\subsection{Monte-Carlo-Like algorithm} 

\begin{figure}[h]
\centering
\includegraphics[width=2.5in]{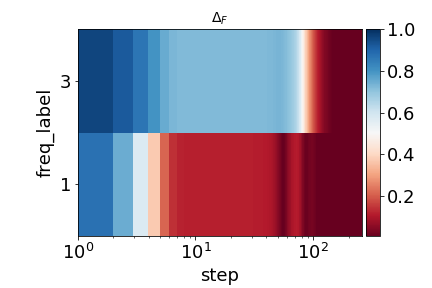}
\caption{Using Monte-Carlo-Like algorithm to learn $f(x)= \sin x  + \sin 3x $. $\Delta_{F}(k)$
of selected important frequencies against different training epochs.}
\label{fig:mc}
\end{figure}

We found the F-Principle also holds in the following Monte-Carlo-Like algorithm: $\bm{\theta}_{j+1}$ is any element in 
$$S=\{{\theta}|L(\theta)<L(\bm{\theta_j}), \left\Vert{\theta-\bm{\theta_j}}\right\Vert<\delta\},$$
unless $S$ is empty. We train a DNN of 1-500-200-1 to learn target function $f(x)= \sin x +\sin 3x$. Considering the important frequency peaks of the target function. As shown in Figure \ref{fig:mc}, the DNN converges the low-frequency components first. 

\section{Discussion and Conclusion} \label{sec:conclusion}

In this paper, we report the F-Principle in the training process of DNN using several non-gradient-descent-based methods. These empirical studies significantly extend the current understanding of the F-Principle. For future work, it is worth to study a mechanism of the F-Principle independent of gradient descent. 


\bibliographystyle{iclr2020_conference}
\bibliography{DLRef}


\newpage
\begin{appendix}

\section{Gradient}
 For the evaluation of gradient, we use the following difference scheme:
 \begin{equation}
\begin{aligned}
L_{\theta_i}(\bm{\theta})&\approx g_{\theta_i}(\bm{\theta})=\frac{L(\cdots, \bm{\theta_{i-1}},\bm{\theta_{i}}+\zeta, \bm{\theta_{i+1}},\cdots)-L(\cdots, \bm{\theta_{i-1}},\bm{\theta_{i}}, \bm{\theta_{i+1}},\cdots)}{\zeta}\\
\nabla L &=(L_{\theta_1},\cdots,L_{\theta_i},\cdots,L_{\theta_P})\approx g_\theta=(g_{\theta_1},\cdots,g_{\theta_i},\cdots,g_{\theta_P})\ \ \ \ \ 
\forall\ 0\le i\le P
\end{aligned}
\end{equation}
Here, $P=(m_{0}+1)\times m_{1}+(m_{1}+1)\times m_{2} + (m_{2}+1)\times m_{3}+\cdots(m_{H-1}+1)\times m_{H}$ is the number of the parameters, and $\zeta$ is a small number defined by user, usually associated with $\epsilon$ or the accuracy of the machine's floating point: $acc$. We set $\zeta=\sqrt{acc}$, which is approximately $1.49\times10^{-8}$ in the following experiments. 

\section{1-d search}
Before the introduction of the 1-d line search, there are some sub-options needed to introduce first, the evaluation of $\alpha_0$ is in the Algorithm  1:





\begin{algorithm} 
\caption{Evaluation of $\alpha_0$ in 1-d search}
\begin{algorithmic}[1]

\Procedure{ }{Input $\alpha_{\min}$, $\alpha_{\max}$, $\alpha_r$, $\phi$, $i$} 

\If{$i>0$}
    \State $cchk:=0.2\left\vert\alpha_{\max}-\alpha_{\min}\right\vert,\alpha_0:=cubicmin(\alpha_{\max},\alpha_{\min},\alpha_r,\phi)$
\EndIf 

\If{$i>0$  or $min\{\left\vert\alpha_{\max}-\alpha_0\right\vert,\left\vert\alpha_{\min}-\alpha_0\right\vert\}\le cchk$}
    \State $qchk:=0.1\left\vert\alpha_{\max}-\alpha_{\min}\right\vert,\ \alpha_0:=\min(\alpha_{\max},\alpha_{\min},\phi)$
    \Else
    \State output $\alpha_0$ and end  this algorithm 
\EndIf 

\If{$\min\{\left\vert\alpha_{\max}-\alpha_0\right\vert,\left\vert\alpha_{\min}-\alpha_0\right\vert\}\le qchk$}
    \State  $\alpha_0:=\frac{\alpha_{\max}+\alpha_{\min}}{2}$, output $\alpha_0$ and end this algorithm 
\EndIf
\EndProcedure
\end{algorithmic}
\end{algorithm}


In the algorithm, $cubicmin(\alpha_{\max},\alpha_{\min},\alpha_r,\phi)$ returns the minimum of $C(x)$ in $[\alpha_{\min},\alpha_{\max}]$,
where $C(x)$ is a cubic polynomial that goes through the points $(\alpha_{\min},\phi(\alpha_{\min}))$, $(\alpha_{\max},\phi(\alpha_{\max})),\ (\alpha_r,\phi(\alpha_r))$ with derivative at $\alpha_{\min}$ of $\phi'(\alpha_{\min})$; in the $\textbf{Step3}$, $quadmin(\alpha_{\max},\alpha_{\min},\phi)$ returns the minimum of $Q(x)$ in $[\alpha_{\min},\alpha_{\max}]$, where $Q(x)$ is a quadratic polynomial that goes through the points $(\alpha_{\min},\phi(\alpha_{\min}))$, $(\alpha_{\max},\phi(\alpha_{\max}))$ with derivative at $\alpha_{\min}$ of $\phi'(\alpha_{\min})$.

In the 1-d line search, under a given boundary $\alpha_{\max}$, we use the following method (for the computation of $\bm{\theta_{k+1}}$ in our algorithms) to find a point in interval $[\bm{\theta_k},\bm{\theta_k}+\alpha_{\max}\bm{d}]$, which makes the searching process \rd{satisfy} strong Wolfe conditions in Algorithm 2.

\begin{algorithm}
\caption{Finding $\alpha_0$ satisfy strong Wolfe conditions}
\begin{algorithmic}[1]

\Procedure{ }{Input Input $\bm{\theta_k}$, direction $\bm{d}$} 

\State Set $i:=0$, $\alpha_{min}:=0$, $\alpha_{max}:=50$, $\alpha_0=1$, $\alpha_r=0$, $\sigma=0.9$, $\rho=10^{-4}$, $maxiter=10$, $\phi(\alpha)=L(\bm{\theta_{k}}+\alpha\bm{d})$, $\alpha_0=1.01\times2\frac{\phi(\alpha_0)-\phi(0)}{\phi'(0)}$

\If{$\alpha_0<0$ or $\alpha_0=Null$}
    \State $\alpha_0=1$
\EndIf 

\While{$i\leq maxiter$ and $\left\vert\phi'(\alpha_0)\right\vert >-\sigma\phi'(\alpha_{min})$}
    \If{$\phi(\alpha_0)>\phi(0)+\rho\alpha_0\phi'(0)$ or $\phi(\alpha_0)\ge\phi(\alpha_{min})$}
        \State $\alpha_r:=\alpha_{max}$, $\alpha_{max}:=\alpha_0$;
    \ElsIf{$\phi'(\alpha_0)(\alpha_{max}-\alpha_{min})\ge0$}
        \State $\alpha_r:=\alpha_{max}$, $\alpha_{max}:=\alpha_{min}$;
        \State $\alpha_{min}=\alpha_0$;
    \Else
        \State $\alpha_r=\alpha_{min}$;
        \State $\alpha_{min}=\alpha_0$;
    \EndIf
    \State Evaluate $\alpha_0$ using $\alpha_{min}$, $\alpha_{max}$, $\alpha_r$, $i$, $\phi$;
    \State $i:=i+1$
\EndWhile

\If{$\left\vert\phi'(\alpha_0)\right\vert\le-\sigma\phi'(\alpha_{min})$}
    \State Output: $\bm{\theta_{k+1}}:=\bm{\theta_k}+\alpha_0\bm{d}$, and end this algorithm;
\Else
    \State Output: 'algorithm failed' and end this algorithm;
\EndIf

\EndProcedure
\end{algorithmic}
\end{algorithm}


In algorithm 2, there is an option called $'Evaluate\ \alpha_0\ '$, which is the algorithm 1, and $\phi'$ is defined by the difference scheme similar in Appendix A. One may use some other algorithm to make the searching satisfy strong Wolfe conditions, the above one is also used in Python's scipy.optimize.


\begin{algorithm}
\caption{CG}
\begin{algorithmic}[1]

\Procedure{ }{Input $\bm{\theta_j}:=\theta_0$, $\epsilon>0$, $M\in\mathbb{N}$} 

\State Set $j:=0$;

\State Compute $g_j=g(\bm{\theta_j})$;

\While{$j\leq M$ and $\left\Vert{g_j}\right\Vert>\epsilon$}
    \State $\bm{d_j}:=-g_j$;
    \State Do 1-d line search $(\bm{d}:=\bm{d_j})$, evaluate $\bm{\theta_{j+1}}$, $j:=j+1$;
    \State Compute $g_j=g(\bm{\theta_j})$;
    \State $\beta:=\frac{g_j^T(g_j-g_{j-1})}{g_{j-1}^Tg_{j-1}}$, $\bm{d_j}:=-g_j+\beta d_{j-1}$;
    \If{$\bm{d_j}^Tg_j>0$}
        \State $\bm{d_j}:=-g_j$;
    \EndIf
\EndWhile

\State output $\bm{\theta_j}$ and end this algorithm;

\EndProcedure
\end{algorithmic}
\end{algorithm}



\begin{algorithm}
\caption{TNC}
\begin{algorithmic}[1]

\Procedure{ }{Input $\bm{\theta_j}:=\theta_0$, $\epsilon>0$, $M\in\mathbb{N}$} 

\State Set $j:=0$;

\State Compute $g_j=g(\bm{\theta_j})$, $H_j=H(\bm{\theta_j})$;

\While{$j\leq M$ and $\left\Vert{g_j}\right\Vert>\epsilon$}
    \State $p_0=0$, $r_0=-g_j$, $l_0=r_0$, $\delta_0=r_0^Tr_0$, $i=0$;
    \While{$l_i^Tq_i>\epsilon\delta_i$ and $i<M$}
        \State $\alpha_i=\frac{r_i^Tr_i}{l_i^Tq_i}$, $p_{i+1}=p_i+\alpha_il_i$, $r_{i+1}=r_i-\alpha_iq_i$
        \If{$\frac{\left\Vert{r_{i+1}}\right\Vert}{\left\Vert{g_j}\right\Vert}\le\eta$}
            \State $\bm{d_j}=\bm{p_i}$;
            \State break;
        \Else
            \State $\beta_i=\frac{r_{i+1}^Tr_{i+1}}{r_i^Tr_i}$, $l_{i+1}=r_{i+1}+\beta_il_i$, $\delta_{i+1}=r_{i+1}^Tr_{i+1}+\beta_i^2\delta_i$;
            $i:=i+1$;
        \EndIf
    \EndWhile
    
    \State $\bm{d_j}=
                    \left\{
                    \begin{aligned}
                    l_0&\ i=0\\
                    p_i&\ i\ge0
                    \end{aligned}
                    \right.$
                    
    \State Do 1-d line search $(\bm{d}:=\bm{d_j})$, evaluate $\bm{\theta_{j+1}}$, $j:=j+1$;
    \State Compute $g_j=g(\bm{\theta_j})$, $H_j=H(\bm{\theta_j})$;
\EndWhile

\State output $\bm{\theta_j}$ and end this algorithm;

\EndProcedure
\end{algorithmic}
\end{algorithm}




\begin{algorithm}
\caption{BFGS}
\begin{algorithmic}[1]

\Procedure{ }{Input $\bm{\theta_j}:=\theta_0$, $\epsilon>0$, $M\in\mathbb{N}$, $H_0=I$} 

\State Set $j:=0$;

\State Compute $g_j=g(\bm{\theta_j})$;

\While{$j\leq M$ and $\left\Vert{g_j}\right\Vert>\epsilon$}
    \State $\bm{d_j}:=-H_jg_j$;
    \State Do 1-d line search $(\bm{d}:=\bm{d_j})$, evaluate $\bm{\theta_{j+1}}$, $g_{j+1}=g(\bm{\theta_{j+1}})$, $s_j=\bm{\theta_{j+1}}-\bm{\theta_{j}}$, $y_j=g_{j+1}-g_j$;
    \State Compute $H_{j+1}=(I-\frac{s_jy_j^T}{s_j^Ty_j})H_{j}(I-\frac{y_js_j^T}{s_j^Ty_j})+\frac{s_js_j^T}{s_j^Ty_j}$;
    \State $\beta:=\frac{g_j^T(g_j-g_{j-1})}{g_{j-1}^Tg_{j-1}}$, $\bm{d_j}:=-g_j+\beta d_{j-1}$, $j=j+1$;
\EndWhile

\State output $\bm{\theta_j}$ and end this algorithm;

\EndProcedure
\end{algorithmic}
\end{algorithm}



\begin{algorithm}
\caption{L-BFGS}
\begin{algorithmic}[1]

\Procedure{ }{Input $\bm{\theta_j}:=\theta_0$, $\epsilon>0$, $M\in\mathbb{N}$, $m\in\mathbb{N}$} 

\State Set $j:=0$;

\State Compute $g_j=g(\bm{\theta_j})$, $\bm{d_0}=g_0$;

\While{$j\leq M$ and $\left\Vert{g_j}\right\Vert>\epsilon$}
    \State Do 1-d line search $(\bm{d}:=\bm{d_j})$, evaluate $\bm{\theta_{j+1}}$, $g_{j+1}=g(\bm{\theta_{j+1}})$, $s_j=\bm{\theta_{j+1}}-\bm{\theta_{j}}$, $y_j=g_{j+1}-g_j$, $\rho_j=\frac{1}{y_j^Ts_j}$, $j:=j+1$;
    \If{$j\le m$}
        \State $\delta:=0$, $L:=j$;
    \Else
        \State $\delta:=j-m$, $L=m$;
    \EndIf
    \State $q_L=g_j$, $i=L-1$;
    \While{$i\geq0$}
        \State $k=i+\delta$, $\alpha_i=\rho_ks_k^Tq_{i+1}$, $q_i=q_{i+1}-\alpha_iy_k$;
        \State $i:=i-1$;
    \EndWhile
    \State $r_0=Iq_0=q_0$, $i:=0$;
    \While{$i\leq L-1$}
        \State $k=i+\delta$, $\beta_i=\rho_ky_k^Tr_i$, $r_{i+1}=r_i+(\alpha_i-\beta_i)s_k$;
        \State $i:=i+1$;
    \EndWhile
    \State $\bm{d_j}=r_L$;
\EndWhile

\State output $\bm{\theta_j}$ and end this algorithm;

\EndProcedure
\end{algorithmic}
\end{algorithm}



\begin{algorithm}
\caption{Powell's algorithm}
\begin{algorithmic}[1]

\Procedure{ }{Input $\bm{\theta_j}:=\theta_0$, $y_0=\bm{\theta_j}$, $\epsilon>0$, $M\in\mathbb{N}$, $S_0=(s_0,s_1,\cdots,s_{p-1})=I$} 

\State Set $j:=0$, $k:=1$;

\State Compute $g_j=g(\bm{\theta_j})$;

\While{$k\leq p$}
    \State $\lambda_{k-1}=argminL(y_{k-1}+\lambda s_{k-1})$, $y_k=y_{k-1}+\lambda_{k-1}s_{k-1}$;
    \State $k:=k+1$;
\EndWhile
\State $s_p=y_p-y_0$;

\While{$j\leq M$ and $\left\Vert{s_p}\right\Vert>\epsilon$}
    \State $\Delta_m=max\{L(y_{i})-L(y_{i+1})$, $0\le i\le p-1\}=L(y_{m})-L(y_{m+1})$;
    \State $f_1=L(y_0)$, $f_2=L(y_p)$, $f_3=L(2y_p-y_0)$;
    \If{$2(f_1-2f_2+f_3)(f_1-f_2-\Delta_m)^2<\Delta(f_1-f_3)^2$}
        \State $\lambda_p=argminL(y_p+\lambda s_p)$, $\bm{\theta_{j+1}}=y_p+\lambda_ps_p$;
        \State $s_k=s_{k+1}$, for $k=m:p-1$;
    \Else
        \State $\bm{\theta_{j+1}}=y_p$, $j:=j+1$;
    \EndIf
    \While{$k\leq p$}
        \State $\lambda_{k-1}=argminL(y_{k-1}+\lambda s_{k-1})$, $y_k=y_{k-1}+\lambda_{k-1}s_{k-1}$;
        \State $k:=k+1$;
    \EndWhile
    \State $s_p=y_p-y_0$;
\EndWhile

\State output $\bm{\theta_j}$ and end this algorithm;

Note: This algorithm uses golden section method (0.618 method) to find the minimum of $\lambda$ in the bracket between $[0,1]$. 

\EndProcedure
\end{algorithmic}
\end{algorithm}



\begin{algorithm}
\caption{PSO}
\begin{algorithmic}[1]

\Procedure{ }{Input $\bm{\theta_j}:=\theta_0$, $\epsilon>0$, $M\in\mathbb{N}$, $m\in\mathbb{N}$, $N=2p$, $H_0=(I,-I)=(h_1,\cdots,h_N)$, $\bm{\theta_j}^{(1)},\bm{\theta_j}^{(2)},\cdots,\bm{\theta_j}^{(N)}$ are randomly set around $\theta_0$} 

\State Set $j:=0$;

\State Set ${\alpha_j}^{(i)}=\arg \min \{L(\alpha),\alpha \in \{ L(\bm{\theta_0}^{(i)}),L(\bm{\theta_1}^{(i)}),\cdots,L(\bm{\theta_j}^{(i)})\}\},\ \forall1\le i\le N$
\State $\bm{\theta_{j+1}}=\arg \min \{L(\alpha),\alpha \in \{ {\alpha_j}^{(1)},{\alpha_j}^{(2)},\cdots,{\alpha_j}^{(N)}\}\}$;

\While{$j\leq M$ and $\left\Vert{\bm{\theta_{j-m}}-\bm{\theta_{j}}}\right\Vert>\epsilon$}
    \State $\bm{\theta_j}^{(i)}:=\bm{\theta_{j-1}}^{(i)}+h_i+2r_1^{(i,j)}({\alpha^{(i)}_{j}}-\bm{\theta_{j-1}}^{(i)})+2r_2^{(i,j)}(\bm{\theta_{j}}-\bm{\theta_{j-1}}^{(i)})$;
    \State Set ${\alpha_j}^{(i)}=\arg \min \{L(\alpha),\alpha \in \{ L(\bm{\theta_0}^{(i)}),L(\bm{\theta_1}^{(i)}),\cdots,L(\bm{\theta_j}^{(i)})\}\},\ \forall1\le i\le N$
    \State $\bm{\theta_{j+1}}=\arg \min \{L(\alpha),\alpha \in \{ {\alpha_j}^{(1)},{\alpha_j}^{(2)},\cdots,{\alpha_j}^{(N)}\}\}$;
    \State $j:=j+1$;
\EndWhile

\State output $\bm{\theta_j}$ and end this algorithm;

Note: $r_1^{(i,j)},\ r_2^{(i,j)}\sim U([0,1])$, while $U([0,1])$ is the uniform distribution on $[0,1]$.
\EndProcedure
\end{algorithmic}
\end{algorithm}



\begin{algorithm}
\caption{Monte-Carlo-like method}
\begin{algorithmic}[1]

\Procedure{ }{Input $\bm{\theta_j}:=\theta_0$, $\epsilon>0$, $M\in\mathbb{N}$, $m\in\mathbb{N}$} 

\State Set $j:=0$;

\While{$j\leq M$ and $\left\Vert{L(\bm{\theta}_j)-L(\bm{\theta}_{j-m})}\right\Vert>\epsilon$}
    \State $\bm{\theta^{(k)}_j}\sim N(\bm{\theta_j},\delta)$ independently, $\forall\ 1\le k\le M$;
    \State $\bm{\theta_{j+1}}:=argmin\{L(\bm{\theta_j}),L(\bm{\theta_j^{(k)}})|\ \forall\ 1\le k\le M\}$, $j:=j+1$
\EndWhile

\State output $\bm{\theta_j}$ and end this algorithm;

\EndProcedure
\end{algorithmic}
\end{algorithm}


\end{appendix}

\end{document}